\documentclass[11pt]{article}

\usepackage[final]{acl}

\usepackage{times}
\usepackage{latexsym}
\usepackage[T1]{fontenc}
\usepackage[utf8]{inputenc}
\usepackage{microtype}
\usepackage{inconsolata}

\usepackage{graphicx}
\usepackage{amsmath,amssymb,amsfonts}
\usepackage{booktabs}
\usepackage{multirow}
\usepackage{array}
\usepackage{xcolor}
\usepackage{colortbl}
\usepackage[most]{tcolorbox}
\usepackage{xurl}

\title{Better Retrieval, Worse Robustness:\\
How Multi-hop RAG Amplifies Upstream ASR Errors}

\author{Zhenghua Bao\\
  Continuum AI \\
  \texttt{zheng.hua.b@gmail.com}}

\begin{document}
\maketitle

\begin{abstract}
Speech-based applications pass spoken queries through
automatic speech recognition (ASR) before any retrieval
module, so ASR errors enter the pipeline as a fixed upstream constraint. We empirically test whether two
extensions to standard retrieval-augmented generation
(RAG), entity-graph linking and iterative reformulation,
absorb or amplify these errors. Using four English accents
synthesized through neural TTS, we evaluate four RAG
configurations on three multi-hop QA benchmarks (HotpotQA, 2WikiMultiHopQA and MuSiQue) against a
clean-text oracle. Although the structurally richer configurations generally retain higher absolute F1 under ASR input, both extensions amplify the error: the F1 gap from clean text to the highest-WER accent is 36--67\% larger under their combination than under naive dense retrieval, on all three benchmarks. The dominant failure mode is corruption of one or more query entities, accounting for 87--96\% of degradation cases on 2WikiMultiHopQA across all four methods.
Two lightweight surface-form mitigations leave most of
the gap intact, indicating that downstream retrieval
structure amplifies remaining entity errors. We release code and data
at \url{https://github.com/Continuum-AI-Corp/spoken-multihop-rag}.
\end{abstract}

\section{Introduction}

\begin{figure}[!t]
    \centering
    \includegraphics[width=\columnwidth]{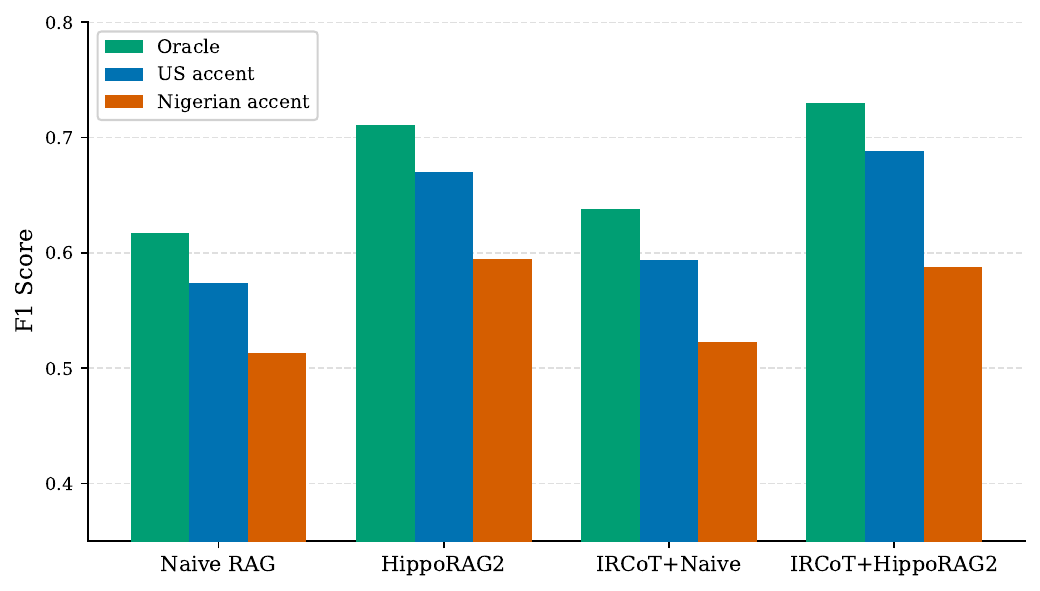}
    \caption{HotpotQA F1 under oracle (clean text),
    US-accented (WER 9.4\%), and Nigerian-accented (WER
    14.5\%) speech. The oracle--NG gap grows from
    0.104 for the simplest baseline (Naive RAG) to 0.142 for
    the most structurally complex configuration
    (IRCoT+HippoRAG2).}
    \label{fig:motivation}
\end{figure}

Voice has become a primary communication channel in everyday
applications. Whether interacting with a voice assistant or
multimodal agent, the user's spoken query is passed
through an automatic speech recognition (ASR) module that
transcribes it into text. However, speakers differ within a language: accents vary widely, and ASR models do not transcribe all variants equally well.
State-of-the-art systems achieve single-digit word error
rates (WER) on clean read-speech benchmarks \citep{whisper},
but their accuracy degrades sharply on accented input.
\citet{koenecke2020racial} report that ASR systems produce nearly $2\times$ higher WER for African American
speakers than for white speakers across five commercial systems, and
\citet{markl2022language} reports similar disparities on Google and Amazon ASR for stigmatized regional varieties of British and Irish English and for second-language speakers.
Figure~\ref{fig:motivation} previews this in our setup: the oracle--NG F1 gap on HotpotQA grows from 0.104 under Naive RAG to 0.142 under IRCoT+HippoRAG2, a 36.5\% larger drop, though the more complex method achieves higher F1 on clean text. These findings motivate our setting, but our study is not a fairness audit. We use \emph{disparity} to denote differences in ASR error rate across our controlled accent conditions, and our claims concern how the resulting errors propagate through retrieval rather than outcomes for specific speaker groups.

The downstream consequences of ASR errors have been studied for machine translation \citep{ruiz2014assessing}, spoken language
understanding \citep{ruan2020towards}, and reading comprehension
\citep{li2018spoken}. Yet the impact of ASR errors on \emph{retrieval-augmented
generation} (RAG) has received little attention. In its standard form, RAG
retrieves passages from an external corpus using the input
query and conditions a generator on the retrieved evidence
\citep{lewis2020retrieval}. This single-step formulation is
already sensitive to query corruption, since retrieval
quality depends directly on the surface form of the query. Throughout, \emph{query corruption} refers to changes that ASR introduces to
query tokens on which retrieval depends, and \emph{query-entity corruption}
to the subset involving named entities: the deletion, substitution, or severe
distortion of a named entity in the transcribed query relative to the
reference text. The problem is compounded in multi-hop QA, which requires
retrieving and reasoning over evidence across multiple
documents \citep{hotpotqa,twowiki,musique}. Recent multi-hop RAG methods extend the standard
pipeline with entity-graph linking
\citep{gutierrez2025rag} or iterative reformulation
\citep{ircot}, each adding a new point at which upstream
ASR errors can propagate. It remains
underexplored whether these methods are more or less robust
to accented input than Naive RAG, and through which
mechanisms they fail.

To analyze whether upstream ASR errors propagate through multi-hop RAG, we ask four research
questions.
\textbf{RQ1:} How much does accented-speech ASR error
degrade multi-hop QA performance, and does it scale with WER?
\textbf{RQ2:} Do structurally complex multi-hop RAG
architectures absorb or amplify these errors?
\textbf{RQ3:} What is the dominant failure mechanism by which
ASR errors propagate through structurally complex RAG?
\textbf{RQ4:} Do lightweight surface-form mitigations close
the gap, and what does the remaining gap reveal about how the failure arises? To answer these, we evaluate four representative RAG
methods on three multi-hop QA benchmarks (HotpotQA,
2WikiMultiHopQA, MuSiQue), using TTS-synthesized spoken
queries across four English accents.

Our experiments reveal that ASR errors propagate through
multi-hop RAG, that structurally complex retrieval methods
amplify rather than absorb these errors, and that entity
corruption is the dominant mechanism. Our main contributions
are:
\begin{itemize}
\item We construct a spoken multi-hop QA evaluation suite
spanning three benchmarks, four English accents, and four
RAG methods, totaling $12{,}000$ spoken
queries\footnote{We release the source code and the data at
\url{https://github.com/Continuum-AI-Corp/spoken-multihop-rag}. The code is released under Apache-2.0, and the
licenses of the underlying datasets and services are documented
in the repository.}.
\item We show that structurally complex retrieval methods
(entity-graph linking, iterative reformulation) amplify
upstream ASR errors despite higher peak F1 on clean text.
\item We identify query-entity corruption as the dominant failure mode across
all four RAG methods (87--96\% on 2WikiMultiHopQA, 67--82\% on HotpotQA,
54--78\% on MuSiQue).
\item Two surface-form mitigations (N-best decoding,
phonetic correction) serve as diagnostic probes. Their
limited recovery indicates downstream retrieval
structure amplifies remaining entity errors. We
additionally validate against $500$ real Nigerian
utterances, supporting TTS as a controlled probe of
the mechanism.
\end{itemize}

\section{Related Work}

\paragraph{Bias and propagation of ASR errors.}
ASR disparities are well studied.
\citet{tatman2017gender} quantified gender and dialect
bias in YouTube auto-captions across two genders and five
dialect groups. \citet{koenecke2020racial} reported nearly $2\times$ higher
WER for African American speakers than white speakers across
five commercial systems. The same pattern appears
across regional varieties of British and Irish English and
second-language speakers
\citep{markl2022language}, and across nine global English
accents \citep{dichristofano2023globalperformancedisparitiesenglishlanguage}. ASR errors also propagate into downstream NLP tasks. \citet{ruiz2014assessing}
showed that MT systems are particularly sensitive to ASR
errors. \citet{ruan2020towards}
demonstrated that intent classification and slot filling F1
drop sharply when entity-bearing tokens are corrupted by
ASR, suggesting that downstream tasks relying on specific
lexical items are particularly vulnerable.
\citet{li2018spoken} extended this observation to reading
comprehension, showing that ASR errors substantially
degrade SQuAD F1. We focus on the propagation mechanism rather than the
upstream disparity: how a fixed ASR error pattern
propagates through multi-hop retrieval architectures,
where inter-document inference creates additional failure points.

\paragraph{Multi-hop retrieval-augmented generation.}
Multi-hop QA benchmarks \citep{hotpotqa,twowiki,musique}
require chaining evidence across multiple documents and pose
distinct challenges from single-hop open-domain QA. Naive
dense retrieval \citep{karpukhin2020dense} provides a strong
baseline but struggles when the bridging entity connecting
sub-questions is not explicitly present in the query. To
address this, HippoRAG \citep{gutierrez2024hipporag} extracts entity-relation
triples from the corpus, builds a knowledge graph linking
documents through shared entities, and retrieves via
personalized PageRank seeded from query entities. HippoRAG2
\citep{gutierrez2025rag} extends this design by additionally
retrieving propositional facts and combining the fact-level
dense-retrieval signal with the PageRank score. In a separate design direction,
IRCoT \citep{ircot} interleaves retrieval with
chain-of-thought reasoning, issuing a new query at each
reasoning step rather than relying on a single retrieval
pass. Both designs introduce new dependencies on the surface
form of the query. \citet{sciavolino2021entity} study this fragility in clean text, showing that dense retrieval fails on entity-centric questions,
with rare entities producing the largest performance drops. ASR
corruption induces a similar effect that compounds across
hops in multi-hop methods.

\paragraph{Spoken question answering.}
Prior work on spoken QA has focused on single-hop
reading comprehension \citep{li2018spoken,lee2018odsqa}
and conversational QA \citep{spokenCoQA}, where
mitigation operates within a single retrieve-and-read
pass per turn.
\citet{faisal2021sdqa} introduced SD-QA, a dialectal spoken
QA benchmark with real recordings across five English
varieties, and report that dialect-induced ASR errors
degrade single-hop QA unevenly across speaker groups. We
extend this line of research to multi-hop reasoning: how
the propagation mechanism changes when the downstream
system is multi-hop retrieval rather than single-hop
reading comprehension. More recent work has built benchmarks for evaluating
large speech-language models on single-hop spoken QA
\citep{wang2025audiobench, chen2026voicebench}, but
multi-hop reasoning over a retrieved corpus is not
tested, and retrieval architectures are rarely
examined. To our knowledge, no prior work
systematically characterizes accent robustness in
multi-hop RAG or isolates the role of entity
corruption in this setting.

\section{Pipeline}

\begin{figure*}[t]
\centering
\includegraphics[width=\textwidth, trim={0 35pt 0 8pt}, clip]{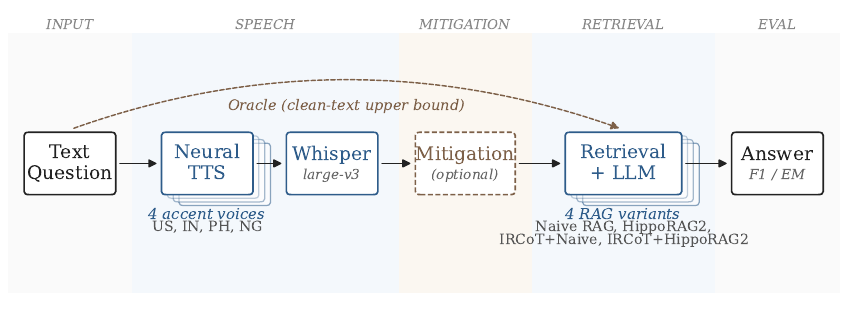}
\caption{Overview of our spoken multi-hop QA evaluation pipeline. Each text
question is synthesized in US, Indian, Filipino, and Nigerian accents through
neural TTS, then transcribed by Whisper-large-v3. The mitigation stage
denotes three separately evaluated experimental conditions (no mitigation,
N-best decoding, or phonetic entity correction) rather than a runtime branch.
Each transcription is then processed by one of the four RAG methods. The
oracle condition (brown dashed arc) bypasses the speech pipeline and feeds
the original text directly to retrieval, providing a clean-text upper bound.}
\label{fig:pipeline}
\end{figure*}

Figure~\ref{fig:pipeline} shows our evaluation pipeline. The pipeline supports configurable combinations of accents, retrieval methods, and mitigation approaches, while holding the question content, retrieval corpus, and generator fixed.

\subsection{Speech Synthesis and Transcription}
Each sampled question is synthesized into speech using four
accented English voices from Microsoft Edge
TTS\footnote{\url{https://learn.microsoft.com/en-us/azure/ai-services/speech-service/}}:
\texttt{en-US-JennyNeural} (US),
\texttt{en-IN-NeerjaNeural} (Indian),
\texttt{en-PH-RosaNeural} (Filipino), and
\texttt{en-NG-EzinneNeural} (Nigerian). All four voices are female and share comparable speaking
style and audio quality, which reduces -- though it does not eliminate -- speaker, prosody, and channel confounds.
The synthesized speech is transcribed with Whisper-large-v3
\citep{whisper} using greedy decoding, yielding four
transcribed queries per question that share the same source
content but may differ in surface form due to ASR error. In
addition to these four accented conditions, we evaluate an
oracle condition that bypasses the speech pipeline and feeds
the original clean text directly to retrieval; the oracle
provides a per-method upper bound that isolates the
degradation from ASR error. We validate the TTS-synthesized
accents against real accented speech in
Section~\ref{sec:real_validation}.

\subsection{RAG Methods}
We evaluate four RAG methods. All methods use
\texttt{gpt-4o-mini}\footnote{\url{https://platform.openai.com/docs/models/gpt-4o-mini}}
as the generator with temperature 0 for reproducibility.

\noindent\textbf{Naive RAG.} Top-$k$ dense retrieval over a
passage index built with
\texttt{text-embedding-3-small}\footnote{\url{https://platform.openai.com/docs/guides/embeddings}};
retrieved passages are concatenated as context for
generation. We use $k=10$, with documents chunked at 400
tokens and 80-token overlap. This serves as the simplest
baseline with no additional graph or iterative dependencies on the query.

\noindent\textbf{HippoRAG2} \citep{gutierrez2025rag}. Extracts
entity-relation triples and propositional facts from the
corpus, builds a knowledge graph linking documents through
shared entities, and retrieves by combining personalized
PageRank seeded from query entities with a fact-level dense
retrieval pass. We use the official implementation with
damping factor 0.5 and \texttt{gpt-4o-mini} for triple and
fact extraction.

\noindent\textbf{IRCoT+Naive} \citep{ircot}. Wraps Naive RAG
in a chain-of-thought loop. At each step, the LLM emits a
reasoning sentence and a follow-up search query; the new
query retrieves additional context that conditions the
next step. The loop terminates when the LLM emits
\texttt{DONE} or after $3$ steps. Retrieved chunks are
deduplicated across steps and capped at $3{,}000$ tokens.

\noindent\textbf{IRCoT+HippoRAG2.} The same iterative loop,
with HippoRAG2 as the underlying retriever. This
configuration combines graph-based linking with iterative
reformulation and achieves the highest F1 on clean text in
our experiments.

\subsection{Mitigation Strategies}
We evaluate two lightweight mitigations applied to the ASR
transcription before retrieval. Both are motivated by the
error analysis in Section~\ref{sec:error}, which attributes
most failures to query-entity corruption. We frame these mitigations as diagnostic
probes: each targets a surface-level form of
entity error, so the remaining gap characterizes what simple corrections cannot recover. Both
are inference-time only, require no retraining, and compose
with any retrieval method. Full algorithmic details and hyperparameters are in
Appendix~\ref{app:mitigations}.

\noindent\textbf{N-best Decoding.} For each spoken query we decode
Whisper once at each of five temperatures
$\tau \in \{0, 0.2, 0.4, 0.6, 0.8\}$, deduplicate the resulting
hypotheses by text, retrieve independently for each hypothesis, and
pass the union of their retrieved documents to the generator
\citep{williams2008nbest,chia2010lattice}. The intuition is that if
the $\tau{=}0$ hypothesis corrupts a critical entity, a
higher-temperature sample may recover the correct form in another
decoding path.

\noindent\textbf{Phonetic Entity Correction.} We replace
ASR-extracted entities with their phonetically nearest
matches from a precomputed corpus entity index, keyed by
Double Metaphone codes and reranked by edit
distance \citep{levenshtein1966binary}. The intuition is
that ASR confusions preserve phonetic structure, so the
correct surface form is often recoverable by phonetic
neighborhood search even when exact string match fails.

\section{Experimental Setup}

\subsection{Datasets}

We evaluate on three widely used multi-hop QA benchmarks.
\textbf{HotpotQA} \citep{hotpotqa} contains bridge and
comparison questions over Wikipedia. \textbf{2WikiMultiHopQA}
\citep{twowiki} adds compositional and inference questions
with explicit reasoning chains. \textbf{MuSiQue}
\citep{musique} spans $2$- to $4$-hop reasoning composed
from single-hop primitives and is regarded as the hardest
of the three. Following prior work \citep{press2023measuring, ircot,
gutierrez2025rag}, we sample $1{,}000$ questions from each
validation set uniformly at random and use the associated
supporting and distractor passages as the retrieval corpus
for each benchmark.

To verify that our synthesized speech produces ASR
error patterns comparable to real English speech, we
additionally transcribe the first $500$ utterances from a
publicly available Nigerian-accented English corpus with
the same Whisper-large-v3 system. This set is used only
for ASR error pattern validation, since the utterances are
general-purpose speech rather than multi-hop QA questions.
Results are reported in Section~\ref{sec:real_validation}.

\subsection{Implementation Details}
Whisper-large-v3 inference runs on a server with an
\texttt{L40S} GPU, while all other components run on a local
machine. We use FAISS \citep{douze2025faiss} with cosine
similarity for the dense index. Embedding generation
(\texttt{text-embedding-3-small}) and answer generation
(\texttt{gpt-4o-mini}) use the OpenAI API. For NER we use
spaCy's \texttt{en\_core\_web\_sm} model. Dense retrieval returns the top $k{=}10$ passages, and IRCoT runs for at
most three reasoning steps, terminating earlier when it produces an answer. Random sampling
uses a fixed seed of $42$.

\subsection{Evaluation Metrics}
Following standard practice on multi-hop QA, we report
token-level F1 and Exact Match (EM) between the generated
answer and the ground-truth answer after SQuAD-style normalization \citep{rajpurkar2016squad}. The evaluated output (the ``Answer'' box in Figure~\ref{fig:pipeline}) is the raw generated
string, and no acceptance test or confidence check is applied after
generation. F1 and EM are
averaged across questions within each dataset-accent cell.
For ASR quality we report Word Error Rate (WER) between the
transcription and the original question text, computed with
Levenshtein-based alignment after lowercasing and
whitespace tokenization. Per-accent F1 gaps are reported as
$\Delta\text{F1} = \text{F1}_{\text{oracle}} - \text{F1}_{\text{accent}}$. We use absolute rather than
relative differences because oracle F1 varies substantially
across datasets. We call a question a degradation case for a given method and accent if the
method answers it correctly from the oracle input but incorrectly from the
ASR transcription, with correctness defined as token-level F1 $\ge 0.5$
against the gold answer. Statistical significance of cross-condition F1 and EM
differences is assessed via paired bootstrap with $10{,}000$
resamples.

\section{Results and Analysis}
\label{sec:results}
\begin{table*}[th]
  \centering
  \small
  \begin{tabular}{llccccc>{\columncolor{gray!15}}c}
    \toprule
    \textbf{Dataset} & \textbf{Method} & \textbf{Oracle}
    & \textbf{US} & \textbf{IN} & \textbf{PH} & \textbf{NG}
    & \textbf{Gap} \\
    \midrule
    & WER & --- & 9.4\% & 11.0\% & 10.6\% & 14.5\% & --- \\
    \cmidrule(lr){2-8}
    \multirow{4}{*}{HotpotQA}
    & Naive RAG    & 0.617 & 0.574 & 0.558 & 0.561 & 0.513 & $0.104$ \\
    & IRCoT+Naive  & 0.638 & 0.594 & 0.573 & 0.583 & 0.523 & $\underline{0.115}$ \\
    & HippoRAG2    & 0.711 & 0.670 & 0.645 & 0.652 & 0.595 & $0.116$ \\
    & IRCoT+HippoRAG2 & 0.730 & 0.688 & 0.652 & 0.671 & 0.588 & $\mathbf{0.142}$ \\
    \midrule
    & WER & --- & 13.6\% & 14.1\% & 13.8\% & 17.1\% & --- \\
    \cmidrule(lr){2-8}
    \multirow{4}{*}{2WikiMultiHopQA}
    & Naive RAG    & 0.468 & 0.399 & 0.378 & 0.393 & 0.331 & $0.137$ \\
    & IRCoT+Naive  & 0.516 & 0.399 & 0.406 & 0.415 & 0.339 & $\underline{0.177}$ \\
    & HippoRAG2    & 0.583 & 0.501 & 0.510 & 0.494 & 0.421 & $0.162$ \\
    & IRCoT+HippoRAG2 & 0.645 & 0.562 & 0.534 & 0.536 & 0.450 & $\mathbf{0.195}$ \\
    \midrule
    & WER & --- & 5.2\% & 5.5\% & 5.7\% & 7.9\% & --- \\
    \cmidrule(lr){2-8}
    \multirow{4}{*}{MuSiQue}
    & Naive RAG    & 0.282 & 0.264 & 0.256 & 0.261 & 0.239 & $0.043$ \\
    & IRCoT+Naive  & 0.306 & 0.267 & 0.260 & 0.264 & 0.243 & $\underline{0.063}$ \\
    & HippoRAG2    & 0.355 & 0.326 & 0.324 & 0.321 & 0.287 & $0.068$ \\
    & IRCoT+HippoRAG2 & 0.381 & 0.339 & 0.344 & 0.346 & 0.309 & $\mathbf{0.072}$ \\
    \bottomrule
  \end{tabular}
  \caption{F1 across three multi-hop QA datasets under
  oracle (clean text), US-accented, Indian-accented (IN),
  Filipino-accented (PH), and Nigerian-accented (NG)
  speech. WER per accent is reported per dataset. Gap
  (shaded) is the oracle--Nigerian F1 difference. Bold
  marks the largest gap per dataset; underline marks
  IRCoT+Naive to isolate the contribution of iterative reformulation. All oracle--Nigerian gaps are significant at
  $p<0.001$.}
  \label{tab:main}
\end{table*}

We organize the section by research question. We find
that (i) ASR errors degrade multi-hop QA performance, with
degradation scaling with WER; (ii) structurally complex
retrieval methods amplify rather than absorb these errors;
(iii) query-entity corruption is the dominant failure mode; and (iv) lightweight surface-form mitigations close
only a small fraction of the gap.

\subsection{ASR Error Degrades Multi-hop QA and Scales with WER (RQ1)}
Table~\ref{tab:main} reports F1 across the four accent
conditions and three benchmarks. Across every (dataset,
method) cell, accented speech reduces F1 relative to the
oracle, and Nigerian-accented input, which carries the
highest WER on all three benchmarks, produces the largest
drop. The oracle--NG gap ranges from $0.043$ (Naive RAG on
MuSiQue) to $0.195$ (IRCoT+HippoRAG2 on 2WikiMultiHopQA),
and is statistically significant at $p<0.001$ across all
cells. Relative to oracle F1, these gaps correspond to
$15$--$34\%$ degradation, with the largest relative drops
on 2WikiMultiHopQA, consistent with its entity-dense
bridge and compositional question format, a pattern we
return to in RQ3. Exact Match results follow the same
pattern: EM gaps track F1 gaps tightly (Pearson
$r > 0.95$ across all $48$ cells) and exhibit the same
method-level amplification ordering on all three datasets
(full EM numbers in Appendix~\ref{app:em}). This answers
the first half of RQ1: ASR error materially degrades
multi-hop QA performance.

\begin{figure}[t]
    \centering
    \includegraphics[width=\columnwidth]{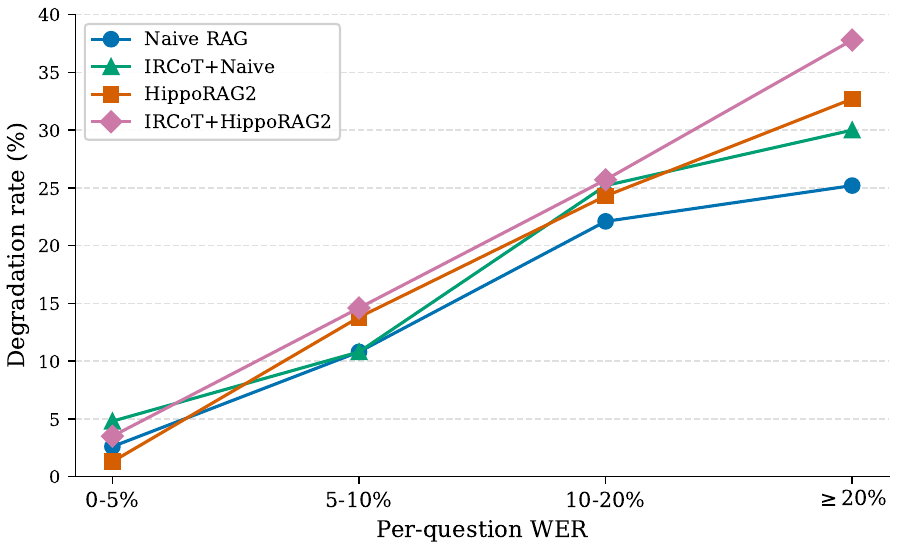}
    \caption{Per-question degradation rate on 2WikiMultiHopQA under Nigerian-accented speech, stratified by Whisper WER.  The gap between Naive RAG and
IRCoT+HippoRAG2 reaches $12.6\%$ at
$\geq 20\%$ WER.}
    \label{fig:wer_threshold}
\end{figure}

The gap also tracks WER. Across accents, WER is lowest for
US and highest for NG on all three benchmarks, and the
oracle--accent F1 gaps in Table~\ref{tab:main} follow the
same ordering in nearly every cell. Across all cells,
mean WER and the oracle--accent F1 gap correlate at
Pearson $r = 0.88$, confirming that higher upstream
WER corresponds to larger downstream degradation. WER
therefore accounts for much of the cross-cell variance; what
aggregate WER misses is which tokens get corrupted, a
question we examine in RQ3. For a finer-grained view,
Figure~\ref{fig:wer_threshold} stratifies questions in the Nigerian-accented
2WikiMultiHopQA condition by individual WER.
Per-question degradation rises monotonically with WER for
all four methods: questions under 5\% WER degrade at rates close to zero, while questions
above 20\% WER degrade in 25.2\% (Naive RAG) to 37.8\%
(IRCoT+HippoRAG2) of cases. This high WER bin is not a long tail: it contains
$n=413$ questions ($41\%$ of the NG pool), the largest of
the four WER bins. Taken with the cross-accent ordering,
this establishes the second half of RQ1: degradation
scales with WER across accents and questions.

\subsection{Structural Complexity Amplifies Rather Than
Absorbs ASR Errors (RQ2)}\label{sec:rq2}

We distinguish robustness from absolute performance: \emph{amplification}
denotes a larger F1 gap between clean text and ASR input within the same
method, not lower absolute F1 under ASR noise. The two can diverge. In the
Nigerian-accented 2WikiMultiHopQA condition, the largest oracle--NG gap
belongs to IRCoT+HippoRAG2 (0.195 vs. 0.137 for Naive RAG), yet the same
method achieves the highest F1 under ASR input (0.450 vs. 0.331). Structural
complexity can therefore raise absolute accuracy and widen the gap at the same time.

The larger gaps in Table~\ref{tab:main} align
with the added structural extensions. We report absolute
F1 differences for direct cross-method comparison within
a dataset, and additionally report relative gap increases to control for the higher
oracle ceiling of the more complex methods. Adding
iterative reformulation alone (IRCoT+Naive vs.\ Naive
RAG) widens the oracle--NG F1 gap by $10.6\%$ on HotpotQA
($0.104 \to 0.115$), $29.2\%$ on 2WikiMultiHopQA
($0.137 \to 0.177$), and $46.5\%$ on MuSiQue
($0.043 \to 0.063$). Adding entity-graph linking alone
(HippoRAG2 vs.\ Naive RAG) widens it by $11.5\%$,
$18.2\%$, and $58.1\%$ respectively. The relative
ordering between these two single-extension
configurations varies by dataset, but combining both
yields the largest gap on every dataset ($36.5\%$,
$42.3\%$, and $67.4\%$ respectively). The combined IRCoT+HippoRAG2 amplification is statistically
significant on all three benchmarks
($p \leq 0.007$).

The combined configuration (IRCoT+HippoRAG2) achieves the
highest oracle F1 on all three benchmarks ($0.730$
HotpotQA, $0.645$ 2WikiMultiHopQA, $0.381$ MuSiQue) yet
produces the largest oracle--NG gap, larger by $36.5\%$
on HotpotQA ($0.104 \to 0.142$), $42.3\%$ on
2WikiMultiHopQA ($0.137 \to 0.195$), and $67.4\%$ on
MuSiQue ($0.043 \to 0.072$) than Naive RAG. The configuration that performs best on clean input shows
the largest drop under ASR noise.
Figure~\ref{fig:wer_threshold} (RQ1) previewed this at
the per-question level: as WER increases, per-question failure rates grow faster for graph-based and iterative methods than for Naive RAG, showing the same pattern.

\begin{figure}[t]
    \centering
    \includegraphics[width=\columnwidth]{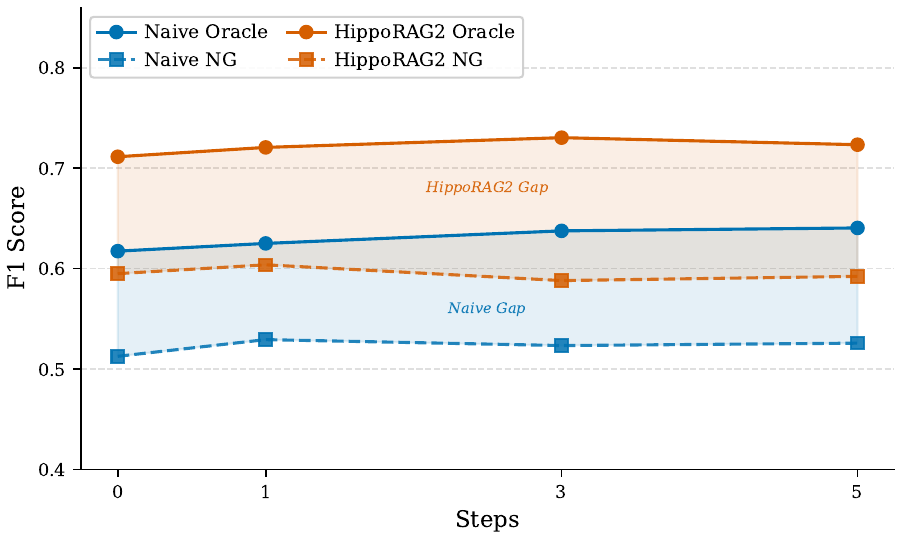}
    \caption{F1 of Naive RAG and HippoRAG2 across IRCoT
steps on $1{,}000$ HotpotQA questions. Solid lines show
oracle F1, dashed lines show Nigerian-accented (NG) F1,
and shaded regions show the oracle--NG gap.}
    \label{fig:step}
\end{figure}

To understand which iterative steps contribute most to
the amplification, we vary the IRCoT step count on
HotpotQA and report the oracle--NG gap at each step
(Figure~\ref{fig:step}). Step $0$ corresponds to the
baselines without IRCoT. The pattern is non-monotone. At
step $1$, the gap closes slightly on Naive RAG
($0.104 \to 0.096$) and has little impact on HippoRAG2
($0.116 \to 0.117$). A single iterative round therefore
does not amplify the error. For Naive RAG it slightly
helps: retrieved context can surface the correct entity,
which the LLM uses in its follow-up query. At step $3$,
the gap reaches $0.115$ on Naive RAG and peaks at $0.142$
on HippoRAG2. Beyond step $3$, the two retrievers
diverge. Naive RAG saturates, with the gap holding at
$0.115$ at step $5$ and oracle and NG F1 both drifting up
by $0.003$. HippoRAG2's gap narrows to $0.131$, but this
reflects degradation rather than recovery: oracle F1
drops by $0.007$ from step $3$ to step $5$ while NG rises
by only $0.004$, meaning the upper bound falls toward the
corrupted condition rather than the other way around.
Even after this partial closure, HippoRAG2's step-$5$ gap
($0.131$) remains $14\%$ larger than Naive RAG's
($0.115$): the structural amplification persists.

This answers RQ2: structural extensions amplify rather
than absorb ASR error, and the iterative component
contributes through cumulative reformulation steps
over successive retrieval rounds.

\subsection{Entity Corruption is the Dominant Failure
Mechanism (RQ3)}\label{sec:error}
\begin{table}[t]
\centering
\small
\setlength{\tabcolsep}{4pt}
\resizebox{\columnwidth}{!}{%
\begin{tabular}{lcccc}
\toprule
\textbf{Error type} & \textbf{US} & \textbf{IN} & \textbf{PH} & \textbf{NG} \\
\midrule
Entity corruption    & 90\% & 87\% & 91\% & 94\% \\
Severe garbling      & 30\% & 35\% & 37\% & 40\% \\
Number/date corrupt. & 4\%  & 4\%  & 4\%  & 4\%  \\
Function-word noise  & 1\%  & 0\%  & 0\%  & 0\%  \\
Other content change & 7\%  & 7\%  & 5\%  & 3\%  \\
\midrule
\# degradation cases & 113  & 124  & 116  & 174  \\
\bottomrule
\end{tabular}%
}
\caption{Distribution of error types within degradation
cases on 2WikiMultiHopQA under Naive RAG. Cases can
carry multiple labels.}
\label{tab:error_types}
\end{table}

Error categorization is automated and rule-based.
For each question we compute a word-level difference
between the original text and the ASR transcription using
Python's \texttt{difflib.SequenceMatcher}\footnote{%
\url{https://docs.python.org/3/library/difflib.html}}.
We classify each non-equal operation by token type:
a capitalized multi-word span (entity proxy),
a digit token (number proxy), or a function word from
a fixed list. Per-question labels (e.g., \emph{entity
corruption}, \emph{severe garbling}) are assigned
based on these per-operation counts and the question's
overall WER. Labels are not mutually exclusive. Rules,
thresholds, and the function-word list are in
Appendix~\ref{app:error_labels}.

Table~\ref{tab:error_types} reports the distribution on
2WikiMultiHopQA under Naive RAG. Entity corruption is the
dominant category for every accent ($87\%$--$94\%$ of
cases), and severe garbling appears in $30\%$--$40\%$.
Numerical, function-word, and miscellaneous edits
are individually small ($\le 7\%$). The Nigerian accent has the highest rates in both
dominant categories ($94\%$ entity corruption,
$40\%$ severe garbling): at higher WER, only the frequency of the dominant
mechanism rises, not the mechanism itself. On HotpotQA and MuSiQue, entity corruption under Naive RAG is less frequent but remains the largest
category, at $67\%$--$82\%$ across accents on HotpotQA
and $59\%$--$78\%$ on MuSiQue
(Appendix~\ref{app:error_types_full}). The effect is strongest on 2WikiMultiHopQA, where
bridge and comparison questions chain multiple named
entities by design.

\begin{figure}[t]
\begin{tcolorbox}[
  colback=gray!4,
  colframe=gray!35,
  boxrule=0.5pt,
  arc=3pt,
  left=5pt, right=5pt, top=4pt, bottom=4pt
]
\small
\textbf{Q:} Which film has the director born earlier,
\textbf{\textcolor[RGB]{0,114,178}{Hanuman Patal Vijay}}
or Young And Dangerous: The Prequel?

\smallskip
\textbf{US:} \ldots\
\textbf{\textcolor[RGB]{255,165,0}{Hanuman Patil Vijay}}
or Young and \ldots\

\smallskip
\textbf{NG:} \ldots\
\textbf{\textcolor[RGB]{213,94,0}{Honourable Peter Vijay}}
or Young and \ldots\

\tcblower

\centering\small
\begin{tabular}{lccc}
\toprule
\textbf{Method} & \textbf{Oracle} & \textbf{US} & \textbf{NG} \\
\midrule
Naive RAG
  & \cellcolor[RGB]{233,245,239}1.00
  & \cellcolor[RGB]{233,245,239}1.00
  & \cellcolor[RGB]{252,235,228}\textbf{\underline{0.00}} \\
HippoRAG2
  & \cellcolor[RGB]{233,245,239}1.00
  & \cellcolor[RGB]{233,245,239}1.00
  & \cellcolor[RGB]{255,248,230}0.50  \\
IRCoT+Naive
  & \cellcolor[RGB]{233,245,239}1.00
  & \cellcolor[RGB]{255,248,230}\textit{0.67}
  & \cellcolor[RGB]{252,235,228}\textbf{\underline{0.00}} \\
IRCoT+HippoRAG2
  & \cellcolor[RGB]{233,245,239}1.00
  & \cellcolor[RGB]{255,248,230}\textit{0.67}
  & \cellcolor[RGB]{252,235,228}\textbf{\underline{0.00}} \\
\bottomrule
\end{tabular}
\end{tcolorbox}

\caption{Case study on 2WikiMultiHopQA illustrating how
a single entity corruption (\emph{Hanuman Patal Vijay}) affects all four RAG methods. Cell values are F1 scores.}
\label{fig:case}
\end{figure}

Figure~\ref{fig:case} illustrates how a single entity
corruption manifests across the four RAG methods. A
2WikiMultiHopQA query about the bridging entity
\emph{Hanuman Patal Vijay} is mildly corrupted under
the US accent (\emph{Hanuman Patil Vijay}) and severely
corrupted under the NG accent (\emph{Honourable Peter
Vijay}). Under NG, Naive RAG fails entirely: the
surrounding tokens (\emph{film}, \emph{director born
earlier}, \emph{Young and Dangerous}) dominate the dense
retrieval signal, and the generator answers with the
comparison film. HippoRAG2 partially recovers: the
uncorrupted \emph{Vijay} token still anchors the entity
graph to the correct region, and the generator produces
\emph{Vijay} for partial credit. The two IRCoT variants fail completely. IRCoT+Naive offers no recovery, since the underlying
retriever has no signal for the iterative loop to build
on. IRCoT+HippoRAG2 is the more revealing case: the
reformulation step anchors on \emph{Honourable Peter
Vijay}, abandons the remaining \emph{Vijay} signal used by the base retriever,
and returns the wrong answer. Even the mild
US corruption (a single-letter change,
\emph{Patal}~$\to$~\emph{Patil}) breaks both IRCoT
variants while the base retrievers stay intact.

To test whether the pattern in Table~\ref{tab:error_types} is specific to the cases that degrade
under Naive RAG, we repeat the same rule-based categorization on the
degradation cases of each of the four configurations (Appendix~\ref{app:error_types_full}). Entity
corruption remains the most frequent category for every method on all three
benchmarks, accounting for 67--82\% of degradation cases on HotpotQA,
87--96\% on 2WikiMultiHopQA, and 54--78\% on MuSiQue. The dominance of entity
corruption is therefore not specific to Naive RAG.

This answers RQ3: corruption of query entities is the dominant failure mode
across configurations and benchmarks, and structural extensions can, as
Figure~\ref{fig:case} illustrates, erase remaining signal preserved by the underlying
retriever during later reformulation and linking steps.

\subsection{Lightweight Mitigations Reveal a Structural Gap (RQ4)}

\begin{table}[t]
\centering
\small
\setlength{\tabcolsep}{4pt}
\begin{tabular}{lcccc}
\toprule
& \multicolumn{2}{c}{\textbf{N-best}}
& \multicolumn{2}{c}{\textbf{Phonetic}} \\
\cmidrule(lr){2-3}\cmidrule(lr){4-5}
\textbf{Method} & F1 & Rec. & F1 & Rec. \\
\midrule
Naive RAG       & 0.328 & $-2.2\%$ & 0.337 & $+4.4\%$  \\
HippoRAG2       & 0.425 & $+2.5\%$ & 0.439 & $+11.1\%$ \\
IRCoT+Naive     & 0.337 & $-1.1\%$ & 0.347 & $+4.5\%$  \\
IRCoT+HippoRAG2 & 0.448 & $-1.0\%$ & 0.464 & $+7.2\%$  \\
\bottomrule
\end{tabular}
\caption{Mitigation results on 2WikiMultiHopQA NG (vs.\
Table~\ref{tab:main} baselines). Recovery is the fraction
of the Oracle--NG gap closed by each mitigation.}
\label{tab:mitigation}
\end{table}

We apply both surface-form mitigations as diagnostic
probes on the cell with the largest degradation (2WikiMultiHopQA
under NG). Each targets a specific hypothesis about the failure. N-best decoding tests whether the gap is reducible to Whisper's sampling noise:
if so, another hypothesis from the same model
should recover the correct entity. Phonetic entity
correction tests whether the gap is reducible to
surface-level entity confusion: if so, the correct
surface form should be recoverable from the corpus by
phonetic neighborhood search. Algorithmic details
and hyperparameters are in
Appendix~\ref{app:mitigations}.

Table~\ref{tab:mitigation} reports F1 and recovery rate
for both mitigations. N-best decoding closes essentially
none of the gap: recovery ranges from $-2.2\%$ (Naive
RAG) to $+2.5\%$ (HippoRAG2), with three of four methods
showing slightly negative recovery. The $\tau{=}0$ hypothesis is
therefore not a noisy sample around a recoverable correct
transcription: pooling hypotheses across five decoding temperatures
yields the same entity corruption. Phonetic
correction does better but still modestly: recovery
ranges from $+4.4\%$ to $+11.1\%$, with HippoRAG2 the
largest beneficiary. The graph-based method's
dependence on exact entity match makes it the most
sensitive to surface-form correction, yet even at this
peak phonetic correction closes only about $11\%$ of
the oracle--NG gap. Neither mitigation closes more than 12\% of the gap in any configuration.

This answers RQ4: lightweight surface-form mitigations
close at most a small fraction of the gap. Targeting
these specific error modes does not close the gap,
indicating that downstream retrieval structure amplifies
remaining entity errors beyond what simple surface
corrections recover.

\subsection{Beyond Synthetic Speech and Whisper}
\label{sec:validation}

To probe sensitivity to two artifacts of our experimental
setup, we validate the ASR error pattern against real
Nigerian speech and rerun the synthesized NG audio with
an alternative ASR system.

\noindent\textbf{Real-speech validation.}
\label{sec:real_validation}
We transcribed $500$ real Nigerian-accented utterances
from a corpus of general-purpose
Nigerian speech\footnote{%
\url{https://huggingface.co/datasets/benjaminogbonna/nigerian_accented_english_dataset}}
with Whisper-large-v3. Real WER reaches $28.9\%$ on
average (median $22.0\%$), $1.7\times$
higher than the $17.1\%$ on our synthesized
2WikiMultiHopQA speech, and at least one entity is
mistranscribed in $51.8\%$ of the utterances that contain a named entity. Synthesized
speech therefore does not overstate ASR noise for real Nigerian speech, and the rate of
entity corruption is comparable across synthetic and
real Nigerian speech
(Appendix~\ref{app:realworld}).

\noindent\textbf{ASR system sensitivity.}
\label{sec:asr_sensitivity}
We re-transcribed the 2WikiMultiHopQA NG
audio with SeamlessM4T-v2-large \citep{barrault2023seamless} and re-ran all
four RAG configurations. Despite a 64\% higher WER (28.0\% vs. 17.1\%), the
ordering of the oracle--ASR gaps is preserved under both systems (Naive RAG
$<$ HippoRAG2 $<$ IRCoT+Naive $<$ IRCoT+HippoRAG2), with the largest gap
growing from 0.195 under Whisper to 0.214 under SeamlessM4T
(Appendix~\ref{app:asr_sensitivity}). The amplification ordering therefore does not appear to be specific to Whisper.

Together, these two checks support the entity-corruption mechanism that
drives our main amplification finding. Neither constitutes end-to-end
evaluation on real spoken multi-hop questions, which we leave to future work.
\section{Conclusion}

We analyzed how upstream ASR errors propagate through
four multi-hop RAG architectures across three benchmarks
and four English accents. ASR errors in accented speech degrade multi-hop QA performance, and degradation
tracks upstream WER (RQ1). Structurally complex retrieval methods amplify rather than
absorb these errors (RQ2). Combining entity-graph linking and iterative reformulation widens the
oracle--NG F1 gap by 36--67\% relative to naive dense retrieval on all three
benchmarks. The richer configurations generally remain more accurate under
ASR input but lose a larger fraction of their clean-text performance. Entity
corruption is the dominant failure mechanism across all four methods, accounting for 87--96\% of degradation cases on 2WikiMultiHopQA and remaining the largest
category on HotpotQA and MuSiQue (RQ3).
Lightweight surface-form mitigations close at most a
small fraction of the gap, suggesting that the failure is
not explained by random ASR noise or surface-level
entity confusion alone (RQ4). The entity-corruption rate is consistent with real Nigerian-accented speech,
and the gap ordering across methods is not driven by the specific ASR system. These results show that structurally complex multi-hop RAG
methods remain vulnerable to upstream ASR noise, leaving
robust multi-hop retrieval under spoken input an open challenge
for future voice-driven systems.

\newpage
\section*{Limitations}

Our evaluation uses synthesized speech rather than
recorded human speech, with a single voice per accent. Beyond Whisper-large-v3, we also evaluate all four configurations with
SeamlessM4T-v2-large on the condition with the largest gap (2WikiMultiHopQA
NG) and recover the same gap ordering across the four methods. This check
remains limited to one dataset and one synthetic voice condition. The $500$ real Nigerian utterances in
Section~\ref{sec:real_validation} show a comparable
entity-corruption rate, but testing additional voices and
speaker attributes, and evaluating on a corpus of real
spoken multi-hop QA queries would strengthen
generalizability. The evaluation is
restricted to English. Code-switched and tonal languages
may exhibit different ASR error patterns. The
error-type analysis in Section~\ref{sec:error} uses a
rule-based proxy for named-entity corruption, and human
validation would strengthen confidence in this
attribution. We use a single LLM
(\texttt{gpt-4o-mini}) to generate answers, and larger
or open-source LLMs may handle corrupted input
differently, particularly inside IRCoT. Promising
directions include LLM-based ASR correction
\citep{ma2023can}, extending correction to every IRCoT
step, falling back to dense retrieval under low ASR
confidence, and evaluation on real accented-speech
corpora such as Common Voice \citep{ardila2020common}
and AfriSpeech-200 \citep{olatunji2023afrispeech}.

\section*{Ethical Considerations}

\noindent\textbf{Accent-related fairness.} In real deployments, speakers with different accents face ASR error
rates that can differ sharply from those of speakers whose varieties
are better represented in ASR training data
\citep{koenecke2020racial,markl2022language}. Our
amplification finding implies that complex multi-hop RAG
methods propagate these upstream errors more than naive
dense retrieval in voice-driven deployments. We flag this as a fairness implication but do not claim
a fairness result. Our speech is synthesized, and the
real-speech validation establishes only a comparable
per-entity corruption rate. It does not establish that
end-to-end RAG behavior on real speech matches the
synthetic setting. Concrete fairness claims require multi-hop QA evaluation
on real spoken queries, which is left for future work.

\noindent\textbf{Synthetic speech and group
representation.} We use a single TTS voice per accent.
This is sufficient as a controlled setting but does not
represent within-accent variation in any speaker group.
Findings should not be read as quantitative claims about
any demographic population. We avoid demographic labels
in our analysis and frame all accent comparisons as
controlled error conditions, not as estimates about
specific speaker groups.

\noindent\textbf{Risk of misuse.} The mechanism we
identify (entity corruption breaks graph-based retrieval)
is a diagnostic finding, not an attack design. The mitigations we benchmark are public, lightweight, and computationally
modest. We do not anticipate dual-use
concerns beyond those general to RAG and ASR research,
such as biased or inaccurate outputs in voice-driven
applications.

\bibliography{custom}

@inproceedings{faisal2021sdqa,
  title={SD-QA: Spoken dialectal question answering for the real world},
  author={Faisal, Fahim and Keshava, Sharlina and Alam, Md Mahfuz Ibn and Anastasopoulos, Antonios},
  booktitle={Findings of the Association for Computational Linguistics: EMNLP 2021},
  pages={3296--3315},
  year={2021}
}

@inproceedings{sciavolino2021entity,
  title={Simple entity-centric questions challenge dense retrievers},
  author={Sciavolino, Christopher and Zhong, Zexuan and Lee, Jinhyuk and Chen, Danqi},
  booktitle={Proceedings of the 2021 Conference on Empirical Methods in Natural Language Processing},
  pages={6138--6148},
  year={2021}
}

@article{koenecke2020racial,
  title={Racial disparities in automated speech recognition},
  author={Koenecke, Allison and Nam, Andrew and Lake, Emily and Nudell, Joe and Quartey, Minnie and Mengesha, Zion and Toups, Connor and Rickford, John R and Jurafsky, Dan and Goel, Sharad},
  journal={Proceedings of the national academy of sciences},
  volume={117},
  number={14},
  pages={7684--7689},
  year={2020},
  publisher={National Academy of Sciences}
}

@inproceedings{markl2022language,
  title={Language variation and algorithmic bias: understanding algorithmic bias in British English automatic speech recognition},
  author={Markl, Nina},
  booktitle={Proceedings of the 2022 ACM Conference on Fairness, Accountability, and Transparency},
  pages={521--534},
  year={2022}
}

@inproceedings{ruiz2014assessing,
  title={Assessing the impact of speech recognition errors on machine translation quality},
  author={Ruiz, Nicholas and Federico, Marcello},
  booktitle={Proceedings of the 11th Conference of the Association for Machine Translation in the Americas: MT Researchers Track},
  pages={261--274},
  year={2014}
}

@inproceedings{ircot,
  title={Interleaving retrieval with chain-of-thought reasoning for knowledge-intensive multi-step questions},
  author={Trivedi, Harsh and Balasubramanian, Niranjan and Khot, Tushar and Sabharwal, Ashish},
  booktitle={Proceedings of the 61st annual meeting of the association for computational linguistics (volume 1: long papers)},
  pages={10014--10037},
  year={2023}
}

@inproceedings{karpukhin2020dense,
  title={Dense passage retrieval for open-domain question answering},
  author={Karpukhin, Vladimir and Oguz, Barlas and Min, Sewon and Lewis, Patrick and Wu, Ledell and Edunov, Sergey and Chen, Danqi and Yih, Wen-tau},
  booktitle={Proceedings of the 2020 conference on empirical methods in natural language processing (EMNLP)},
  pages={6769--6781},
  year={2020}
}

@inproceedings{whisper,
  title={Robust speech recognition via large-scale weak supervision},
  author={Radford, Alec and Kim, Jong Wook and Xu, Tao and Brockman, Greg and McLeavey, Christine and Sutskever, Ilya},
  booktitle={International conference on machine learning},
  pages={28492--28518},
  year={2023},
  organization={PMLR}
}

@inproceedings{hotpotqa,
  title={HotpotQA: A dataset for diverse, explainable multi-hop question answering},
  author={Yang, Zhilin and Qi, Peng and Zhang, Saizheng and Bengio, Yoshua and Cohen, William and Salakhutdinov, Ruslan and Manning, Christopher D},
  booktitle={Proceedings of the 2018 conference on empirical methods in natural language processing},
  pages={2369--2380},
  year={2018}
}

@inproceedings{twowiki,
  title={Constructing a multi-hop qa dataset for comprehensive evaluation of reasoning steps},
  author={Ho, Xanh and Nguyen, Anh-Khoa Duong and Sugawara, Saku and Aizawa, Akiko},
  booktitle={Proceedings of the 28th International Conference on Computational Linguistics},
  pages={6609--6625},
  year={2020}
}

@article{musique,
  title={MuSiQue: Multihop Questions via Single-hop Question Composition},
  author={Trivedi, Harsh and Balasubramanian, Niranjan and Khot, Tushar and Sabharwal, Ashish},
  journal={Transactions of the Association for Computational Linguistics},
  volume={10},
  pages={539--554},
  year={2022},
  publisher={MIT Press One Broadway, 12th Floor, Cambridge, Massachusetts 02142, USA~…}
}

@inproceedings{rajpurkar2016squad,
  title={Squad: 100,000+ questions for machine comprehension of text},
  author={Rajpurkar, Pranav and Zhang, Jian and Lopyrev, Konstantin and Liang, Percy},
  booktitle={Proceedings of the 2016 conference on empirical methods in natural language processing},
  pages={2383--2392},
  year={2016}
}

@article{spokenCoQA,
  title={Towards data distillation for end-to-end spoken conversational question answering},
  author={You, Chenyu and Chen, Nuo and Liu, Fenglin and Yang, Dongchao and Zou, Yuexian},
  journal={arXiv preprint arXiv:2010.08923},
  year={2020}
}

@inproceedings{tatman2017gender,
  title={Gender and dialect bias in YouTube’s automatic captions},
  author={Tatman, Rachael},
  booktitle={Proceedings of the first ACL workshop on ethics in natural language processing},
  pages={53--59},
  year={2017}
}

@inproceedings{lee2018odsqa,
  title={ODSQA: Open-domain spoken question answering dataset},
  author={Lee, Chia-Hsuan and Wang, Shang-Ming and Chang, Huan-Cheng and Lee, Hung-Yi},
  booktitle={2018 IEEE Spoken Language Technology Workshop (SLT)},
  pages={949--956},
  year={2018},
  organization={IEEE}
}

@article{chen2026voicebench,
  title={Voicebench: Benchmarking llm-based voice assistants},
  author={Chen, Yiming and Yue, Xianghu and Zhang, Chen and Gao, Xiaoxue and Tan, Robby T and Li, Haizhou},
  journal={Transactions of the Association for Computational Linguistics},
  volume={14},
  pages={378--398},
  year={2026},
  publisher={MIT Press 255 Main Street, 9th Floor, Cambridge, Massachusetts 02142, USA~…}
}

@inproceedings{williams2008nbest,
  title={Exploiting the ASR n-best by tracking multiple dialog state hypotheses.},
  author={Williams, Jason D},
  booktitle={Interspeech},
  pages={191--194},
  year={2008}
}

@inproceedings{chia2010lattice,
  title={A lattice-based approach to query-by-example spoken document retrieval},
  author={Chia, Tee Kiah and Sim, Khe Chai and Li, Haizhou and Ng, Hwee Tou},
  booktitle={Proceedings of the 31st annual international ACM SIGIR conference on Research and development in information retrieval},
  pages={363--370},
  year={2008}
}

@article{ma2023can,
  title={Can generative large language models perform asr error correction?},
  author={Ma, Rao and Qian, Mengjie and Manakul, Potsawee and Gales, Mark and Knill, Kate},
  journal={arXiv preprint arXiv:2307.04172},
  year={2023}
}

@inproceedings{ardila2020common,
  title={Common voice: A massively-multilingual speech corpus},
  author={Ardila, Rosana and Branson, Megan and Davis, Kelly and Kohler, Michael and Meyer, Josh and Henretty, Michael and Morais, Reuben and Saunders, Lindsay and Tyers, Francis and Weber, Gregor},
  booktitle={Proceedings of the twelfth language resources and evaluation conference},
  pages={4218--4222},
  year={2020}
}

@article{olatunji2023afrispeech,
  title={Afrispeech-200: Pan-african accented speech dataset for clinical and general domain asr},
  author={Olatunji, Tobi  and
      Afonja, Tejumade  and
      Yadavalli, Aditya  and
      Emezue, Chris Chinenye  and
      Singh, Sahib  and
      Dossou, Bonaventure F. P.  and
      Osuchukwu, Joanne  and
      Osei, Salomey  and
      Tonja, Atnafu Lambebo  and
      Etori, Naome  and
      Mbataku, Clinton},
  journal={Transactions of the Association for Computational Linguistics},
  volume={11},
  pages={1669--1685},
  year={2023},
  publisher={MIT Press One Broadway, 12th Floor, Cambridge, Massachusetts 02142, USA~…}
}

@article{levenshtein1966binary,
  author = {Levenshtein, Vladimir I},
  journal = {Soviet Physics Doklady},
  month = {February},
  pages={707--710},
  title = {Binary Codes Capable of Correcting Deletions, Insertions and Reversals},
  volume={10},
  year = 1966
}

@article{gutierrez2025rag,
  title={From rag to memory: Non-parametric continual learning for large language models},
  author={Guti{\'e}rrez, Bernal Jim{\'e}nez and Shu, Yiheng and Qi, Weijian and Zhou, Sizhe and Su, Yu},
  journal={arXiv preprint arXiv:2502.14802},
  year={2025}
}

@article{lewis2020retrieval,
  title={Retrieval-augmented generation for knowledge-intensive nlp tasks},
  author={Patrick Lewis and Ethan Perez and Aleksandra Piktus and Fabio Petroni and Vladimir Karpukhin and Naman Goyal and Heinrich Küttler and Mike Lewis and Yih, Wen-tau and Tim Rocktäschel and Sebastian Riedel and Douwe Kiela},
  journal={Advances in neural information processing systems},
  volume={33},
  pages={9459--9474},
  year={2020}
}

@inproceedings{press2023measuring,
  title={Measuring and narrowing the compositionality gap in language models},
  author={Press, Ofir and Zhang, Muru and Min, Sewon and Schmidt, Ludwig and Smith, Noah A and Lewis, Mike},
  booktitle={Findings of the Association for Computational Linguistics: EMNLP 2023},
  pages={5687--5711},
  year={2023}
}

@article{douze2025faiss,
  title={The faiss library},
  author={Douze, Matthijs and Guzhva, Alexandr and Deng, Chengqi and Johnson, Jeff and Szilvasy, Gergely and Mazar{\'e}, Pierre-Emmanuel and Lomeli, Maria and Hosseini, Lucas and J{\'e}gou, Herv{\'e}},
  journal={IEEE Transactions on Big Data},
  year={2025},
  publisher={IEEE}
}

@article{gutierrez2024hipporag,
  title={Hipporag: Neurobiologically inspired long-term memory for large language models},
  author={Guti{\'e}rrez, Bernal Jim{\'e}nez and Shu, Yiheng and Gu, Yu and Yasunaga, Michihiro and Su, Yu},
  journal={Advances in neural information processing systems},
  volume={37},
  pages={59532--59569},
  year={2024}
}

@article{barrault2023seamless,
  title={Seamless: Multilingual Expressive and Streaming Speech Translation},
  author={{Seamless Communication} and Loïc Barrault and Yu-An Chung and Mariano Coria Meglioli and David Dale and Ning Dong and Mark Duppenthaler and Paul-Ambroise Duquenne and Brian Ellis and Hady Elsahar and Justin Haaheim and John Hoffman and Min-Jae Hwang and Hirofumi Inaguma and Christopher Klaiber and Ilia Kulikov and Pengwei Li and Daniel Licht and Jean Maillard and Ruslan Mavlyutov and Alice Rakotoarison and Kaushik Ram Sadagopan and Abinesh Ramakrishnan and Tuan Tran and Guillaume Wenzek and Yilin Yang and Ethan Ye and Ivan Evtimov and Pierre Fernandez and Cynthia Gao and Prangthip Hansanti and Elahe Kalbassi and Amanda Kallet and Artyom Kozhevnikov and Gabriel Mejia Gonzalez and Robin San Roman and Christophe Touret and Corinne Wong and Carleigh Wood and Bokai Yu and Pierre Andrews and Can Balioglu and Peng-Jen Chen and Marta R. Costa-jussà and Maha Elbayad and Hongyu Gong and Francisco Guzmán and Kevin Heffernan and Somya Jain and Justine Kao and Ann Lee and Xutai Ma and Alex Mourachko and Benjamin Peloquin and Juan Pino and Sravya Popuri and Christophe Ropers and Safiyyah Saleem and Holger Schwenk and Anna Sun and Paden Tomasello and Changhan Wang and Jeff Wang and Skyler Wang and Mary Williamson},
  journal={arXiv preprint arXiv:2312.05187},
  year={2023}
}

@inproceedings{ruan2020towards,
  title={Towards an ASR error robust spoken language understanding system},
  author={Ruan, Weitong and Nechaev, Yaroslav and Chen, Luoxin and Su, Chengwei and Kiss, Imre},
  booktitle={Proc. Interspeech 2020},
  pages={901--905},
  year={2020}
}

@article{li2018spoken,
  title={Spoken squad: A study of mitigating the impact of speech recognition errors on listening comprehension},
  author={Li, Chia-Hsuan and Wu, Szu-Lin and Liu, Chi-Liang and Lee, Hung-yi},
  journal={arXiv preprint arXiv:1804.00320},
  year={2018}
}

@article{dichristofano2023globalperformancedisparitiesenglishlanguage,
  title={Global performance disparities between English-language accents in automatic speech recognition},
  author={DiChristofano, Alex and Shuster, Henry and Chandra, Shefali and Patwari, Neal},
  journal={arXiv preprint arXiv:2208.01157},
  year={2022}
}

@inproceedings{wang2025audiobench,
  title={Audiobench: A universal benchmark for audio large language models},
  author={Wang, Bin and Zou, Xunlong and Lin, Geyu and Sun, Shuo and Liu, Zhuohan and Zhang, Wenyu and Liu, Zhengyuan and Aw, AiTi and Chen, Nancy},
  booktitle={Proceedings of the 2025 Conference of the Nations of the Americas Chapter of the Association for Computational Linguistics: Human Language Technologies (Volume 1: Long Papers)},
  pages={4297--4316},
  year={2025}
}

\newpage
\appendix

\section{Mitigation Details}
\label{app:mitigations}
\subsection{N-best Decoding}
\label{app:nbest}

For each spoken query we decode five hypotheses at decoding temperatures
$\tau \in \{0, 0.2, 0.4, 0.6, 0.8\}$. $\tau=0$ is identical to the greedy decode used without any mitigations. The four sampled hypotheses provide
slightly different surface forms drawn from regions of higher acoustic
uncertainty. Each hypothesis is independently passed to the
retriever, yielding up to five sets of retrieved documents
per query. We pass the union of these document sets to the
generator as context, deduplicated by document ID and
capped at the per-method context budget. For IRCoT-based methods,
N-best is applied only to the initial ASR transcription, 
subsequent IRCoT-generated queries pass through
unchanged. The generator and all other pipeline components
are unchanged.

\subsection{Phonetic Entity Correction}
\label{app:phonetic}

\paragraph{Offline corpus indexing.}
The corpus entity index combines two sources: (i) all
document titles in the retrieval corpus, and (ii) named
entities extracted from passage bodies using spaCy
NER\footnote{Model: \texttt{en\_core\_web\_sm}; entity
types retained: \texttt{PERSON, ORG, GPE, LOC, FAC,
WORK\_OF\_ART, EVENT, PRODUCT, NORP}.}. Each entity string
is normalized and encoded. The index is a mapping from
phonetic-code tuples to canonical entity strings, held in
memory.

\paragraph{Query-time correction.}
Given an ASR transcription, the procedure proceeds as
follows:

\begin{enumerate}
    \item \textbf{NER extraction.} Run spaCy NER on the
    transcription and collect mentioned entities.

    \item \textbf{Phonetic filtering.} Encode each
    candidate with the same Double Metaphone procedure. Retrieve corpus entities whose
    phonetic-code Jaccard similarity with the candidate
    exceeds $0.4$, and retain the top $50$.

    \item \textbf{Edit-distance reranking.} Score each
    retained corpus entity against the candidate using
    \texttt{rapidfuzz.fuzz.ratio}, a normalized Indel similarity on a $0$--$100$ scale.

    \item \textbf{Threshold gating.} If the top match
    exceeds $75$, accept it. Otherwise, fall back to a
    full-corpus search and accept matches only above
    $85$. The stricter fallback threshold protects against
    spurious string matches when phonetic similarity is no
    longer available as a prior.

    \item \textbf{Replacement.} Replace the original
    candidate span in the transcription with the accepted
    canonical form, using word-boundary regex matching to
    avoid corrupting adjacent tokens.
\end{enumerate}

\paragraph{Hyperparameter selection.}
The thresholds (Jaccard $0.4$, edit-distance $75$,
fallback edit-distance $85$) and the top-$K=50$ candidate
pool were chosen by inspection of representative ASR
error cases prior to the main evaluation. They are kept
fixed across all datasets and accents. No tuning is
performed on the $1{,}000$ evaluation samples.

\section{Error Categorization Rules}
\label{app:error_labels}

\paragraph{Pipeline.} For each (question, ASR transcription)
pair we compute a word-level difference using Python's
\texttt{difflib.SequenceMatcher} after
whitespace tokenization, no lowercasing, to preserve case for
entity detection.

\paragraph{Per-operation labels.} A replace or delete operation is labelled based on the original word. An insert operation has no original word, so it matches none of the tests
below and is counted as a content-word change:
\begin{itemize}
  \item \textbf{entity corruption} if the word matches the
  capitalized-phrase regex
  \texttt{\textbackslash b[A-Z][a-z]+(?:\textbackslash s+[A-Z][a-z]+)*\textbackslash b};
  \item \textbf{number corruption} if the word matches
  \texttt{\textbackslash b\textbackslash d+(?:st|nd|rd|th)?\textbackslash b};
  \item \textbf{function-word change} if the word is in our function-word list (Table~\ref{tab:funcwords});
  \item \textbf{content-word change} otherwise.
\end{itemize}

\paragraph{Aggregate labels.} A question receives severe garbling if its WER exceeds $0.20$ and it has
at least one content-word change. It receives
function-word noise if it has function-word changes
and no other label has occurred. It receives other
content change when changes occurred but no
other aggregate label was triggered. Labels are not mutually
exclusive: \textit{entity corruption} and \textit{severe
garbling} frequently co-occur on Nigerian speech.

\begin{table}[t]
\centering
\small
\setlength{\tabcolsep}{4pt}
\begin{tabular}{ll}
\toprule
\textbf{Category} & \textbf{Words} \\
\midrule
Articles (3)        & a, an, the \\
Prepositions (13)   & of, in, on, at, to, for, with, by,\\
                    & from, as, into, about, between \\
Forms of \textit{be} (7) & is, was, are, were, be, been, being \\
Other auxiliaries (6)    & have, has, had, do, does, did \\
Modals (8)               & will, would, shall, should,\\
                         & may, might, can, could \\
Conjunctions (7)         & and, or, but, not, no, if, than \\
Pronouns / det. (10)     & that, which, who, whom, this,\\
                         & these, those, it, its, whose \\
Wh-words (5)             & what, where, when, how, why \\
\bottomrule
\end{tabular}
\caption{Function-word list used for ASR error
categorization. Matching is case-insensitive after stripping non-alphanumeric characters.}
\label{tab:funcwords}
\end{table}

\section{Exact Match Results}
\label{app:em}
\begin{table*}[t]
  \centering
  \small
  \begin{tabular}{llccccc>{\columncolor{gray!15}}c}
    \toprule
    \textbf{Dataset} & \textbf{Method} & \textbf{Oracle}
    & \textbf{US} & \textbf{IN} & \textbf{PH} & \textbf{NG}
    & \textbf{Gap} \\
    \midrule
    & WER & --- & 9.4\% & 11.0\% & 10.6\% & 14.5\% & --- \\
    \cmidrule(lr){2-8}
    \multirow{4}{*}{HotpotQA}
    & Naive RAG       & 0.488 & 0.456 & 0.439 & 0.446 & 0.403 & $0.085$ \\
    & IRCoT+Naive     & 0.508 & 0.476 & 0.453 & 0.469 & 0.417 & $\underline{0.091}$ \\
    & HippoRAG2       & 0.568 & 0.534 & 0.514 & 0.519 & 0.473 & $0.095$ \\
    & IRCoT+HippoRAG2 & 0.588 & 0.546 & 0.519 & 0.534 & 0.463 & $\mathbf{0.125}$ \\
    \midrule
    & WER & --- & 13.6\% & 14.1\% & 13.8\% & 17.1\% & --- \\
    \cmidrule(lr){2-8}
    \multirow{4}{*}{2WikiMultiHopQA}
    & Naive RAG       & 0.419 & 0.346 & 0.330 & 0.343 & 0.286 & $0.133$ \\
    & IRCoT+Naive     & 0.475 & 0.354 & 0.359 & 0.366 & 0.293 & $\underline{0.182}$ \\
    & HippoRAG2       & 0.510 & 0.425 & 0.432 & 0.420 & 0.347 & $0.163$ \\
    & IRCoT+HippoRAG2 & 0.569 & 0.486 & 0.451 & 0.459 & 0.380 & $\mathbf{0.189}$ \\
    \midrule
    & WER & --- & 5.2\% & 5.5\% & 5.7\% & 7.9\% & --- \\
    \cmidrule(lr){2-8}
    \multirow{4}{*}{MuSiQue}
    & Naive RAG       & 0.180 & 0.160 & 0.151 & 0.156 & 0.138 & $0.042$ \\
    & IRCoT+Naive     & 0.209 & 0.172 & 0.166 & 0.174 & 0.158 & $\underline{0.051}$ \\
    & HippoRAG2       & 0.242 & 0.218 & 0.212 & 0.212 & 0.188 & $0.054$ \\
    & IRCoT+HippoRAG2 & 0.270 & 0.239 & 0.240 & 0.246 & 0.215 & $\mathbf{0.055}$ \\
    \bottomrule
  \end{tabular}
  \caption{Exact Match (EM) across three multi-hop QA
  datasets under oracle and four English
  accents. All settings follow the convention in Table~\ref{tab:main}.}
  \label{tab:em_main}
\end{table*}

Table~\ref{tab:em_main} reports Exact Match (EM) for
every condition in Table~\ref{tab:main}. EM is a
stricter metric than F1, requiring exact string match
after normalization. EM gaps are correspondingly smaller
in absolute terms but track F1 gaps tightly, with
Pearson $r > 0.95$ across all cells. The method-level
amplification ordering observed for F1 is preserved on
every dataset, and the largest EM gap reaches $0.189$ on
2WikiMultiHopQA under IRCoT+HippoRAG2, mirroring the
structural amplification pattern reported for F1.

\section{Error-Type Distributions Across Benchmarks and Methods}
\label{app:error_types_full}

\begin{table}[t]
\centering
\setlength{\tabcolsep}{4pt}
\begin{tabular}{lcccc}
\toprule
\textbf{Error type} & \textbf{US} & \textbf{IN} & \textbf{PH} & \textbf{NG} \\
\midrule
Entity corruption    & 67\% & 70\% & 75\% & 82\% \\
Severe garbling      & 14\% & 24\% & 24\% & 44\% \\
Number/date corrupt. & 5\%  & 5\%  & 5\%  & 4\%  \\
Function-word noise  & 4\%  & 9\%  & 8\%  & 7\%  \\
Other content change & 21\% & 13\% & 11\% & 7\%  \\
\midrule
\# degradation cases & 81   & 103  & 88   & 136  \\
\bottomrule
\end{tabular}
\caption{Distribution of error types within degradation
cases on HotpotQA under Naive RAG. Cases can carry
multiple labels.}
\label{tab:error_types_hotpotqa}
\end{table}

\begin{table}[t]
\centering
\setlength{\tabcolsep}{4pt}
\begin{tabular}{lcccc}
\toprule
\textbf{Error type} & \textbf{US} & \textbf{IN} & \textbf{PH} & \textbf{NG} \\
\midrule
Entity corruption    & 59\% & 78\% & 66\% & 68\% \\
Severe garbling      & 4\%  & 7\%  & 14\% & 21\% \\
Number/date corrupt. & 0\%  & 0\%  & 0\%  & 1\%  \\
Function-word noise  & 9\%  & 4\%  & 2\%  & 1\%  \\
Other content change & 28\% & 13\% & 20\% & 23\% \\
\midrule
\# degradation cases & 46   & 54   & 50   & 71   \\
\bottomrule
\end{tabular}
\caption{Distribution of error types within degradation
cases on MuSiQue under Naive RAG. Cases can carry
multiple labels.}
\label{tab:error_types_musique}
\end{table}

\paragraph{Cross-benchmark analysis.} Table~\ref{tab:error_types} reports the error analysis on
2WikiMultiHopQA only, where each question references multiple
named entities. This makes entity corruption the most
visible failure mode. We apply the same rule-based
labeling to HotpotQA and MuSiQue degradation cases under
Naive RAG (Tables~\ref{tab:error_types_hotpotqa} and
\ref{tab:error_types_musique}). Entity corruption remains
the largest single failure category on every accent
across both datasets, confirming that our finding
generalizes beyond 2WikiMultiHopQA.

Two patterns are worth noting. Severe garbling on
HotpotQA NG reaches $44\%$, co-occurring with entity
corruption rather than competing with it. NG questions
are partially destroyed at the sentence level rather
than losing just one entity. ``Other content change''
on MuSiQue US reaches $28\%$, the only cell where a
non-entity category meaningfully competes with entity
corruption. This likely reflects MuSiQue's small number
of degradation cases ($n=46$) and low overall WER
($5.2\%$).

\paragraph{Cross-method analysis.} To verify that the dominance of entity corruption is not specific to Naive
RAG, we repeat the same rule-based categorization on the degradation cases of
each of the four RAG configurations, on all three benchmarks. Table~\ref{tab:cross_method_entity} reports the entity-corruption rate within
each method's degradation cases, with the number of cases in parentheses.
Entity corruption remains the most frequent category in every cell:
67--82\% on HotpotQA, 87--96\% on 2WikiMultiHopQA, and 54--78\% on MuSiQue.

\begin{table*}[t]
  \centering
  \small
  \begin{tabular}{llcccc}
    \toprule
    \textbf{Dataset} & \textbf{Method} & \textbf{US} & \textbf{IN} & \textbf{PH} & \textbf{NG} \\
    \midrule
    \multirow{4}{*}{HotpotQA}
    & Naive RAG       & 67\% (81)  & 70\% (103) & 75\% (88)  & 82\% (136) \\
    & IRCoT+Naive     & 77\% (81)  & 77\% (107) & 73\% (100) & 79\% (154) \\
    & HippoRAG2       & 73\% (71)  & 77\% (96)  & 81\% (90)  & 82\% (154) \\
    & IRCoT+HippoRAG2 & 67\% (75)  & 78\% (110) & 69\% (99)  & 80\% (177) \\
    \midrule
    \multirow{4}{*}{2WikiMultiHopQA}
    & Naive RAG       & 90\% (113) & 87\% (124) & 91\% (116) & 94\% (174) \\
    & IRCoT+Naive     & 95\% (140) & 92\% (132) & 92\% (132) & 94\% (206) \\
    & HippoRAG2       & 92\% (130) & 93\% (120) & 94\% (135) & 94\% (211) \\
    & IRCoT+HippoRAG2 & 95\% (132) & 94\% (156) & 96\% (156) & 94\% (241) \\
    \midrule
    \multirow{4}{*}{MuSiQue}
    & Naive RAG       & 59\% (46)  & 78\% (54)  & 66\% (50)  & 68\% (71)  \\
    & IRCoT+Naive     & 54\% (70)  & 69\% (84)  & 65\% (71)  & 72\% (99)  \\
    & HippoRAG2       & 64\% (61)  & 66\% (65)  & 75\% (69)  & 78\% (108) \\
    & IRCoT+HippoRAG2 & 57\% (81)  & 65\% (78)  & 73\% (73)  & 71\% (112) \\
    \bottomrule
  \end{tabular}
  \caption{Entity-corruption rate within degradation cases for each RAG method
and accent, with the number of degradation cases in parentheses. Cases can
carry multiple labels.}
  \label{tab:cross_method_entity}
\end{table*}

\section{Real-World Speech Validation}
\label{app:realworld}

\paragraph{Dataset.}
We sample the first $500$ utterances from a publicly available corpus on HuggingFace\footnote{%
\url{https://huggingface.co/datasets/benjaminogbonna/nigerian_accented_english_dataset}}. The corpus contains short
Nigerian-accented English utterances from multiple
speakers and is released for research use. Because the utterances are general-purpose speech rather than multi-hop
QA queries, we use the set only for ASR error-pattern validation.

\paragraph{Transcription and WER.}
We transcribe each utterance with the same Whisper-large-v3
configuration used in the main pipeline. Real Nigerian
speech is substantially noisier than our synthesized
Nigerian condition: mean WER on real speech reaches
$28.9\%$ (median $22.0\%$), compared to $17.1\%$ on
2WikiMultiHopQA, $14.5\%$ on HotpotQA, and $7.9\%$ on
MuSiQue under synthesized speech. Our synthesized condition
therefore underestimates the actual ASR difficulty by
a factor of $1.7\times$ to $3.7\times$ across the three
benchmarks. The WER distribution on real speech also has
a heavy tail: a noticeable fraction of utterances exceed
$50\%$ WER, reflecting variation in recording channel
and speaker that TTS cannot reproduce.

\paragraph{Entity corruption rate.}
For each utterance we extract named entities from the
reference text using spaCy NER and check whether each
entity is preserved verbatim in the transcription, after
lowercasing and removing non-alphanumeric characters.
Among the $282$ real utterances containing at least one
entity, $51.8\%$ have at least one mistranscribed entity,
and $40.7\%$ of the $607$ entity instances are altered.
On our synthesized Nigerian 2WikiMultiHopQA condition, 43.8\% of entity
instances are altered, a factor of 1.08$\times$ the real-speech rate. This supports the use of TTS as a
controlled probe for the entity-corruption mechanism
behind the amplification reported in
Section~\ref{sec:rq2}.

\section{ASR Sensitivity Analysis}
\label{app:asr_sensitivity} 

\paragraph{Setup.}
We replace Whisper-large-v3 with SeamlessM4T-v2-large
\citep{barrault2023seamless}, using greedy decoding. All other
components of the pipeline remain unchanged: TTS voices,
question samples, retrieval corpus, RAG methods, and
generator. We rerun the Nigerian-accent condition with all four RAG configurations on
2WikiMultiHopQA, the dataset that exhibited the largest oracle--NG gaps in
our main results.

\begin{table}[t]
\centering
\small
\begin{tabular}{lcccc}
\toprule
& \multicolumn{2}{c}{\textbf{Whisper}} & \multicolumn{2}{c}{\textbf{SeamlessM4T}} \\
\cmidrule(lr){2-3} \cmidrule(lr){4-5}
\textbf{Method} & \textbf{F1} & \textbf{Gap} & \textbf{F1} & \textbf{Gap} \\
\midrule
Naive RAG        & 0.331 & 0.137 & 0.333 & 0.135 \\
IRCoT+Naive      & 0.339 & 0.177 & 0.334 & 0.182 \\
HippoRAG2        & 0.421 & 0.162 & 0.421 & 0.162 \\
IRCoT+HippoRAG2  & 0.450 & 0.195 & 0.431 & 0.214 \\
\bottomrule
\end{tabular}
\caption{End-to-end F1 and oracle--ASR gap on 2WikiMultiHopQA NG under
Whisper-large-v3 (WER 17.1\%) and SeamlessM4T-v2-large (WER 28.0\%). Per-method
oracle F1 is reported in Table~\ref{tab:main}. The gap ordering across methods
is preserved under both ASR systems.}
\label{tab:asr_sensitivity}
\end{table}

\paragraph{Results.} SeamlessM4T yields a mean WER of 28.0\% on the same
2WikiMultiHopQA NG audio, 64\% higher than Whisper-large-v3's
17.1\%. Table~\ref{tab:asr_sensitivity} reports F1 under ASR input and the
oracle--ASR gap under both systems. Despite the larger
upstream error, absolute F1 changes by at most 0.019 (IRCoT+HippoRAG2), and
the gap ordering Naive RAG $<$ HippoRAG2 $<$ IRCoT+Naive $<$
IRCoT+HippoRAG2 is preserved under both systems. The largest gap grows from
0.195 under Whisper to 0.214 under SeamlessM4T, so the system with the higher
WER slightly widens rather than reverses the amplification pattern. The
ordering across methods is therefore not driven by the specific ASR system:
the failure mode is sensitive to entity-level surface-form changes in the
corrupted query, which both systems produce despite their different overall
error rates.

\end{document}